\documentclass{article}

\usepackage[dblblindworkshop, final]{neurips_2025}

\usepackage[utf8]{inputenc} 
\usepackage[T1]{fontenc}    
\usepackage{hyperref}       
\usepackage{url}            
\usepackage{booktabs}       
\usepackage{amsfonts}       
\usepackage{nicefrac}       
\usepackage{microtype}      
\usepackage{xcolor} 
\usepackage{amsmath}
\usepackage{multicol}
\usepackage{multirow}
\usepackage{graphicx}
\usepackage{booktabs}
 \usepackage[ruled]{algorithm2e}
\usepackage{algorithmic}
\usepackage{amssymb}
\usepackage{graphicx}
\usepackage{enumitem}
\usepackage{threeparttable}
\usepackage{float}

\title{EWC-Guided Diffusion Replay for Exemplar Free Continual Learning in Medical Imaging}

\author{
  Anoushka Harit\textsuperscript{1} \\
  \texttt{ah2415@cam.ac.uk} 
  \And
  William Prew\textsuperscript{1} \\
  \texttt{wp287@cam.ac.uk}
  \And
  Zhongtian Sun\textsuperscript{1,2} \\
  \texttt{z.sun-256@kent.ac.uk}
  \And
  Florian Markowetz\textsuperscript{1} \\
  \texttt{fm327@cam.ac.uk}
  \AND
\textsuperscript{1}\textbf{University of Cambridge, Cambridge, United Kingdom} \\
\textsuperscript{2}\textbf{University of Kent, Canterbury, United Kingdom}}

\begin{document}

\maketitle
\begin{abstract}
Medical imaging foundation models must adapt over time, yet full retraining is often blocked by privacy constraints and cost. We present a continual learning framework that avoids storing patient exemplars by pairing class conditional diffusion replay with Elastic Weight Consolidation. Using a compact Vision Transformer backbone, we evaluate across eight MedMNIST v2 tasks and CheXpert. On CheXpert our approach attains 0.851 AUROC, reduces forgetting by more than 30\% relative to DER\texttt{++}, and approaches joint training at 0.869 AUROC, while remaining efficient and privacy preserving. Analyses connect forgetting to two measurable factors: fidelity of replay and Fisher weighted parameter drift, highlighting the complementary roles of replay diffusion and synaptic stability. The results indicate a practical route for scalable, privacy aware continual adaptation of clinical imaging models.
\end{abstract}

\section{Introduction}
Foundation models (FMs) are reshaping medical AI by enabling transferable representations across imaging tasks~\cite{wu2025information}. However, in real-world deployment, they must continually adapt to new diseases, imaging protocols, and workflows. Retraining is costly, and storing patient data for replay is often restricted, making privacy-preserving continual learning essential. This need intensifies \emph{catastrophic forgetting}~\cite{mccloskey1989catastrophic}, where learning new tasks erodes prior knowledge. Existing solutions fall short: regularisation (e.g., EWC~\cite{kirkpatrick2017overcoming}, EFT~\cite{liu2022elastic}) degrades under distribution shift; exemplar-based replay (e.g., DER++~\cite{buzzega2020dark}, SPM~\cite{zhu2021prototype}) violates privacy; generative replay (e.g., VAEs, GANs~\cite{shin2017continual,van2020brain}) struggles with fine medical detail; and dynamic expansion (e.g., PMoE~\cite{jung2024pmoe}, CoPE~\cite{de2021continual}) incurs high compute. \textbf{Our approach.} We propose a hybrid framework that combines a lightweight Vision Transformer (ViT), class-conditional DDPM for exemplar-free replay, and EWC for stability. Diffusion preserves subtle medical detail without storing data, and the unified model handles both 2D and 3D tasks efficiently. We also derive a forgetting bound linking replay fidelity and parameter anchoring. Experiments on MedMNIST v2~\cite{yang2023medmnist} and CheXpert~\cite{irvin2019chexpert} confirm strong retention across modalities ~\cite{sun2022unimodal}.

\textbf{Contributions.}  
(i) A privacy-preserving continual learning framework combining DDPM replay and EWC;  
(ii) unified handling of 2D/3D medical tasks with a single diffusion model;  
(iii) a theoretical bound linking forgetting, replay fidelity, and parameter stability;  
(iv) empirical validation on multi-scale benchmarks.

\section{Continual Learning Setup}
\label{sec:setup}

Let $\mathcal{T}=\{T_1,\dots,T_K\}$ be a sequence of tasks with datasets 
$D_k=\{(x_i^{(k)},y_i^{(k)})\}_{i=1}^{n_k}$, where $x_i^{(k)}\in\mathcal{X}$ and 
$y_i^{(k)}\in\mathcal{Y}_k$. At step $k$, only $D_k$ is available; $D_j$ for $j<k$ cannot be stored (exemplar-free constraint).  

A model $f_\theta:\mathcal{X}\to\Delta(\mathcal{Y})$ with parameters $\theta\in\mathbb{R}^d$ is updated sequentially $\theta^{(k)}\mapsto\theta^{(k+1)}$.  

Forgetting is measured as  
\[
\mathcal{F}=\tfrac{1}{K-1}\sum_{j=1}^{K-1}\max_{t\leq K}\big(A_{j,t}-A_{j,K}\big),
\]
and final accuracy as  
\[
\mathcal{A}=\tfrac{1}{K}\sum_{j=1}^K A_{j,K},
\]
where $A_{j,t}$ is accuracy on $T_j$ after step $t$.  

The objective is to maximise $\mathcal{A}$ while minimising $\mathcal{F}$ under exemplar-free continual learning.

\section{Theoretical Analysis of Forgetting}
\label{theory}
We analyse forgetting in exemplar-free continual learning by decomposing it into two measurable sources that directly motivate our method:  
(i) \emph{distributional drift}, arising from imperfect replay, and  
(ii) \emph{parameter drift}, arising from unstable optimisation.

\begin{figure}[H]
  \centering
  \includegraphics[width=0.52\linewidth]{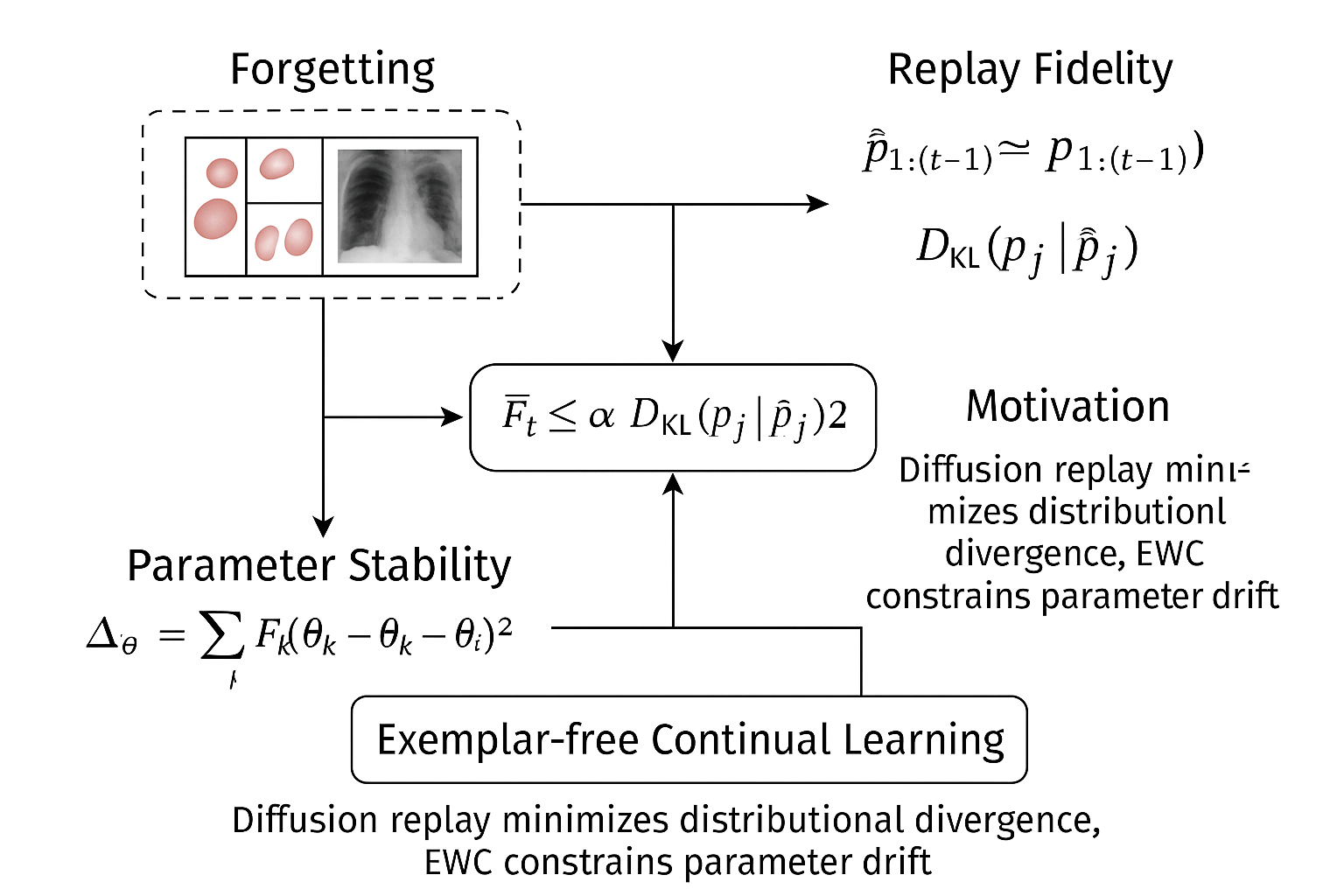}
  \caption{
    Problem formulation of exemplar-free continual learning. 
    Past datasets $\mathcal{D}_{1:(t-1)}$ cannot be stored; the learner sees only the current task $\mathcal{D}_t$ and must perform well across all $\mathcal{D}_{1:T}$. 
    Two drivers of forgetting emerge: replay fidelity (divergence $D_{\mathrm{KL}}(p_j\|\hat p_j)$) and parameter stability (Fisher-weighted drift $\sum_i F_i(\theta_i-\theta_i^\star)^2$).
  }
  \label{fig:problem_formulation}
\end{figure}

\paragraph{Setup.}
For each task $j$, let $\mathcal{A}_j^\star$ denote accuracy immediately after training, and $\mathcal{A}_j^T$ the accuracy after learning all $T$ tasks. Forgetting on task $j$ is
\[
F_j = \mathcal{A}_j^\star - \mathcal{A}_j^T,
\qquad
\bar{F} = \tfrac{1}{T}\sum_{j=1}^T F_j.
\]

\paragraph{Distributional drift.}
When replay substitutes a proxy $\hat{p}_j$ for the true $p_j$, Pinsker’s inequality for bounded loss $\ell \in [0,L_{\max}]$ gives
\[
\Big|\,\mathbb{E}_{p_j}[\ell] - \mathbb{E}_{\hat{p}_j}[\ell]\,\Big|
\;\leq\; L_{\max}\sqrt{\tfrac{1}{2}D_{\mathrm{KL}}(p_j \,\|\, \hat{p}_j)}.
\]
Thus, replay error contributes in proportion to the $\mathrm{KL}$ divergence.  
This term motivates \emph{diffusion replay}, which yields lower divergence than VAEs or GANs, and \emph{Fisher Scheduled Replay}, which allocates generative samples where divergence is most damaging.

\paragraph{Parameter drift.}
Let $\theta_j^\star$ be the optimum for task $j$. A second-order expansion around $\theta_j^\star$ gives
\[
\mathcal{L}(\theta) \approx \mathcal{L}(\theta_j^\star) 
+ \tfrac{1}{2}(\theta - \theta_j^\star)^\top F (\theta - \theta_j^\star),
\]
where $F$ is the Fisher information matrix. The excess loss scales with
\[
D_j = \sum_i F_i(\theta_i - \theta_i^\star)^2,
\]
capturing instability of Fisher-salient parameters.  
This motivates \emph{Elastic Weight Consolidation}, which explicitly penalises this Fisher-weighted drift.

\paragraph{Unified bound.}
Combining the two effects, forgetting can be bounded as
\[
\bar{F} \;\leq\; \alpha \, D_{\mathrm{KL}}(p_j \,\|\, \hat{p}_j)
+ \beta \sum_i F_i(\theta_i - \theta_i^\star)^2,
\]
with constants $\alpha,\beta>0$ depending on loss smoothness and curvature.  
This bound maps directly to our design: diffusion replay reduces the $\mathrm{KL}$ term, FSR further focuses replay where divergence is largest, and EWC constrains the Fisher-weighted drift.

\paragraph{Empirical validation.}
Although $\alpha$ and $\beta$ are not directly observable, both terms of the bound can be estimated. 
For each task $j$ we compute replay divergence $\widehat{D}_{\mathrm{KL}}(p_j \,\|\, \hat{p}_j)$ and Fisher-weighted drift $D_j$, and relate them to observed forgetting via
\[
F_j = a \,\widehat{D}_{\mathrm{KL}}(p_j \,\|\, \hat{p}_j) + b \, D_j + \varepsilon_j.
\]
As reported in Appendix~\ref{derivation_theoretical_forgetting_bound}, both terms correlate positively with forgetting, and the joint regression explains more variance than either alone. This provides empirical support for the replay–drift decomposition and justifies the integrated design of EWC-DR.

\section{Methodology}\label{method}
We propose a \textbf{hybrid continual learning framework} for \emph{privacy-preserving, exemplar-free} adaptation of medical imaging models. It integrates class-conditional DDPM replay to synthesise prior-task data, EWC to preserve critical parameters, and a lightweight ViT backbone to fuse real and replayed samples. Inspired by dual-memory theory~\cite{mcclelland1995there}, DDPM enables fast recall, while EWC supports gradual consolidation.

\begin{figure}[h]
  \centering
  \includegraphics[width=0.55\linewidth]{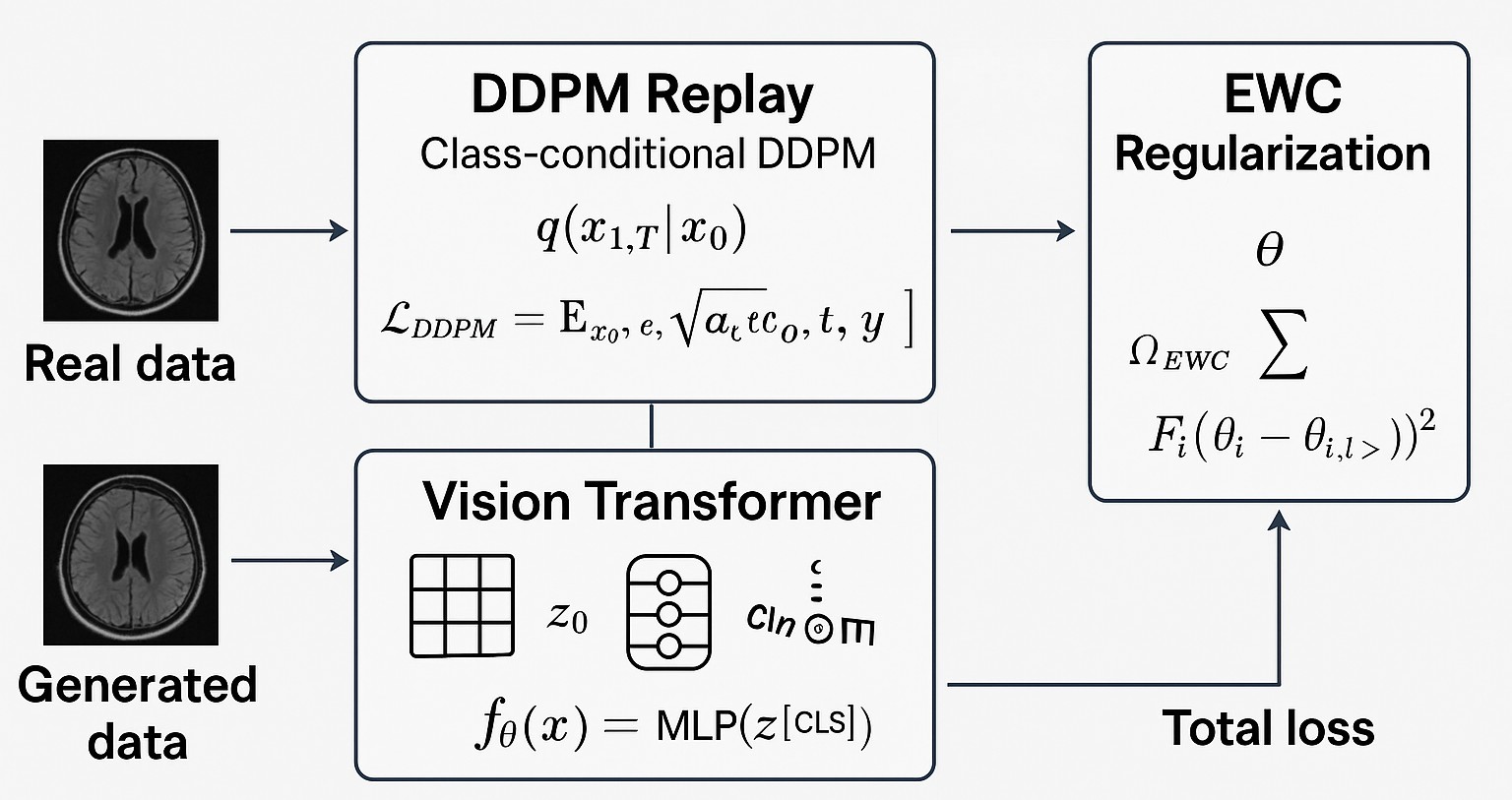}
  \caption{Hybrid continual learning: DDPM replay generates exemplar-free samples, fused with real data in a ViT classifier, with EWC for parameter stability.}
  \label{fig:framework}
\end{figure}

\subsection{Problem Setting}
Let tasks $\{\mathcal{T}_1,\dots,\mathcal{T}_K\}$ have datasets $\mathcal{D}_k=\{(x_i^{(k)},y_i^{(k)})\}_{i=1}^{N_k}$ from evolving, unknown $\mathcal{P}_k(x,y)$. At step $k$, only $\mathcal{D}_k$ is accessible. The goal is to learn $f_\theta$ that generalises to $\mathcal{P}_{\leq k}=\bigcup_{j=1}^k\mathcal{P}_j$ without storing raw past data.

\subsection{Overall Objective}
We minimise:
\begin{equation}
\mathcal{L}_{\mathrm{total}}^{(k)} =
\mathbb{E}_{(x, y) \sim \mathcal{D}_k \cup \hat{\mathcal{D}}_{<k}}
\!\left[ \mathcal{L}_{\mathrm{CE}}(f_\theta(x), y) \right]
+ \lambda \sum_i F_i (\theta_i - \theta^*_{i,<k})^2,
\end{equation}
where $\hat{\mathcal{D}}_{<k}$ are DDPM-generated samples, $F_i$ are Fisher information entries, and $\theta^*_{i,<k}$ are anchored parameters from earlier tasks.

\paragraph{Component 1: DDPM Replay}
For each task $\mathcal{T}_k$, we train a single class-conditional DDPM $p_k(x \mid y)$ to reverse the diffusion process:
\begin{equation}
q(x_{1:T} \mid x_0) = \prod_{t=1}^T \mathcal{N}(x_t; \sqrt{\alpha_t} x_{t-1}, (1 - \alpha_t)\mathbf{I}),
\end{equation}
with training objective:
\begin{equation}
\mathcal{L}_{\mathrm{DDPM}} =
\mathbb{E}_{x_0, y, \epsilon, t} \!
\left\| \epsilon - \epsilon_\phi\!\left(\sqrt{\bar{\alpha}_t}x_0 + \sqrt{1-\bar{\alpha}_t}\epsilon, t, y\right) \right\|^2.
\end{equation}
The generated set $\hat{\mathcal{D}}_k$ is added to the replay buffer for future tasks.

\paragraph{Component 2: ViT Classifier}
An image $x \in \mathbb{R}^{H\times W\times C}$ is split into $N$ patches $x_p^{(i)} \in \mathbb{R}^{P^2C}$:
\begin{equation}
z_0 = [x_p^{(1)}\mathbf{E}; \dots; x_p^{(N)}\mathbf{E}] + \mathbf{E}_{\mathrm{pos}},
\end{equation}
processed through $L$ transformer layers, with [CLS] token $z_L^{[\mathrm{CLS}]}$ classified via:
\begin{equation}
f_\theta(x) = \mathrm{MLP}(z^{[\mathrm{CLS}]}_L).
\end{equation}

\paragraph{Component 3: EWC Regularisation}
EWC penalises drift on important parameters:
\begin{equation}
\Omega_{\mathrm{EWC}} = \sum_i F_i (\theta_i - \theta^*_{i})^2,
\end{equation}
with $F_i$ estimated on past tasks.

\paragraph{Class-Incremental Learning (CIL)}
CIL requires prediction over $\mathcal{C}_{\leq k}$ without task IDs. Our replay buffer maintains class balance, mitigating bias. Eq.~(1) is applied with cross-entropy over the cumulative label set.

\paragraph{Theoretical View: Forgetting Bound}
We bound average forgetting as:\begin{equation}
\bar{F} \leq \alpha \, \mathrm{KL}(\mathcal{P}_k \parallel \hat{\mathcal{P}}_k) +
\beta \, \| \theta_k - \theta^*_{k-1} \|_2^2,
\end{equation}
linking replay fidelity (first term) and parameter stability (second term), motivating joint DDPM replay and EWC.

\section{Experimental Setup and Results}\label{experiments}
\paragraph{Datasets.}
We evaluate on three benchmarks spanning distinct modalities: \textbf{MedMNIST-2D}~\cite{yang2023medmnist} \footnote{https://medmnist.com/} (six 2D tasks: PathMNIST, BloodMNIST, DermaMNIST, RetinaMNIST, BreastMNIST, PneumoniaMNIST), \textbf{MedMNIST-3D} (OrganMNIST3D, NoduleMNIST3D), and \textbf{CheXpert}~\cite{irvin2019chexpert}\footnote{https://aimi.stanford.edu/datasets/chexpert-chest-x-rays} (14-label chest X-ray classification). A class-incremental protocol is used, with disjoint label sets per task and no task identity at test time.

\paragraph{Baselines.}
We compare against regularisation-based methods (\textit{EFT}~\cite{liu2022elastic}, \textit{CoPE}~\cite{de2021continual}) that avoid replay but struggle under distribution shift; replay-based methods (\textit{DER++}~\cite{buzzega2020dark}, \textit{SPM}~\cite{zhu2021prototype}, \textit{VAE+Replay}~\cite{shin2017continual}); and the dynamic-expansion method \textit{PMoE}~\cite{jung2024pmoe}. All baselines use the same ViT backbone and training setup.

\paragraph{Implementation.}
We use a lightweight ViT (patch size 16, 6 layers, 8 heads, hidden dim 512) initialised from ImageNet. 
The DDPM is class-conditional ($T{=}1000$, cosine $\beta_t$) and trained for 200 epochs per task. 
EWC uses Fisher information from 500 samples per class with $\lambda \in \{10, 50, 100\}$ tuned on validation. 
Replay generates 256 class-balanced samples per task, stored as raw float32 tensors within a fixed 100~MB budget (Appendix~\ref{app:budget}). Each batch mixes real and replayed data 1:1. All models use AdamW (lr $3{\times}10^{-4}$, weight decay $0.01$), batch size 64, and results are averaged over five seeds.

\paragraph{Metrics.}
We report \emph{Average Accuracy} after the final task, \emph{Average Forgetting}~\cite{chaudhry2018riemannian}, and \emph{AUC} to reflect clinical relevance in multi-label settings. On CheXpert, we report mean AUC averaged across the 14 labels (macro-average).

\paragraph{Protocols.}
All methods are trained under a fixed canonical task order. Forgetting is measured from intermediate evaluations after each task and reported at the final step. For CheXpert, we report the mean AUC, macro-averaged across the 14 disease labels. Robustness to alternative task orders is analysed separately in Appendix~\ref{app:order}.

\subsection{Main Results}\label{sec:results}
Table~\ref{tab:main_results_all} reports performance across MedMNIST-2D, MedMNIST-3D, and CheXpert. All methods use the same ViT backbone, memory budget (100\,MB), and task sequence for fair comparison. \textbf{DDPM+EWC} consistently achieves the best trade-off between accuracy, AUC, and forgetting, outperforming regularisation-only methods (EWC, CoPE) and replay-based baselines (VAE+Replay, DER++).  
On CheXpert, it reduces forgetting by over 30\% compared to DER++, underscoring the value of high-fidelity generative replay in clinically sensitive settings.  
Joint training remains an upper bound but violates continual learning constraints.\footnote{Finetune denotes sequential training without replay or regularisation and serves as a lower-bound baseline.}

\begin{table}[H]
\centering
\caption{\textbf{Continual Adaptation Results.} Mean over 5 runs. Best in \textbf{bold}, second-best \underline{underlined}. Forgetting ($F\downarrow$) lower is better.}
\label{tab:main_results_all}
\resizebox{\linewidth}{!}{
\begin{tabular}{l|ccc|ccc|ccc}
\toprule
& \multicolumn{3}{c|}{\textbf{MedMNIST-2D}} & \multicolumn{3}{c|}{\textbf{MedMNIST-3D}} & \multicolumn{3}{c}{\textbf{CheXpert}} \\
\textbf{Method} & Acc$\uparrow$ & F$\downarrow$ & AUC$\uparrow$ & Acc$\uparrow$ & F$\downarrow$ & AUC$\uparrow$ & Acc$\uparrow$ & F$\downarrow$ & AUC$\uparrow$ \\
\midrule
Finetune & 67.4 & 27.5 & 0.820 & 63.2 & 29.1 & 0.801 & 64.8 & 26.9 & 0.802 \\
EWC & 72.9 & 19.7 & 0.842 & 68.5 & 21.5 & 0.824 & 70.5 & 19.4 & 0.824 \\
EFT & 71.1 & 21.4 & 0.839 & 67.9 & 22.9 & 0.820 & 69.4 & 20.5 & 0.820 \\
CoPE & 72.4 & 19.9 & 0.843 & 68.8 & 21.3 & 0.826 & 70.8 & 19.2 & 0.826 \\
DER++ & \underline{75.6} & \underline{14.2} & \underline{0.853} & 70.9 & \underline{16.8} & 0.835 & \underline{73.2} & \underline{13.8} & \underline{0.838} \\
SPM & 74.9 & 15.0 & 0.852 & \underline{71.2} & 17.0 & \underline{0.836} & 72.6 & 14.4 & 0.835 \\
VAE+Replay & 74.2 & 15.6 & 0.851 & 70.8 & 17.5 & 0.837 & 71.7 & 15.1 & 0.833 \\
PMoE & 74.5 & 15.3 & 0.852 & 70.9 & 17.2 & 0.835 & 72.1 & 14.7 & 0.834 \\
\textbf{Ours (DDPM+EWC)} & \textbf{78.1} & \textbf{10.5} & \textbf{0.866} & \textbf{74.0} & \textbf{12.9} & \textbf{0.849} & \textbf{76.4} & \textbf{10.9} & \textbf{0.851} \\
Joint (Upper Bound) & 81.4 & 0.0 & 0.879 & 77.5 & 0.0 & 0.861 & 79.1 & 0.0 & 0.869 \\
\bottomrule
\end{tabular}
}
\end{table}

\subsection{Task-wise Performance}
We report accuracy on the first ($T_1$), middle ($T_{\text{mid}}$), and final ($T_n$) tasks after completing the sequence.  
For MedMNIST-2D, $T_1$ and $T_2$ denote datasets (e.g., BreastMNIST, PneumoniaMNIST); for MedMNIST-3D, they refer to OrganMNIST3D and NoduleMNIST3D; and for CheXpert, they correspond to early, median, and final label groups.  
\textbf{DDPM+EWC} consistently preserves $T_1$ performance, particularly in CheXpert where strong domain shift typically causes rapid forgetting (Table~\ref{tab:taskwise_all}).

\begin{table}[h]
\centering
\small
\caption{\textbf{Task-wise accuracy (\%)} after the final task for MedMNIST-2D, MedMNIST-3D, and CheXpert.
Best in \textbf{bold}, second-best \underline{underlined}.}
\label{tab:taskwise_all}
\resizebox{\linewidth}{!}{
\begin{tabular}{l|ccc|ccc|ccc}
\toprule
& \multicolumn{3}{c|}{\textbf{MedMNIST-2D}} & \multicolumn{3}{c|}{\textbf{MedMNIST-3D}} & \multicolumn{3}{c}{\textbf{CheXpert}} \\
\cmidrule(lr){2-4} \cmidrule(lr){5-7} \cmidrule(lr){8-10}
\textbf{Method} & $T_1$ & $T_{\text{mid}}$ & $T_n$ & $T_1$ & $T_{\text{mid}}$ & $T_n$ & $T_1$ & $T_{\text{mid}}$ & $T_n$ \\
\midrule
Finetune     & 45.2 & 60.1 & 83.4 & 41.7 & 56.5 & 79.0 & 43.8 & 58.0 & 80.2 \\
EWC          & 57.5 & 66.3 & 84.1 & 53.2 & 62.9 & 80.5 & 54.9 & 64.2 & 81.0 \\
EFT          & 55.9 & 65.4 & 83.8 & 52.4 & 61.7 & 80.2 & 53.6 & 63.2 & 80.6 \\
CoPE         & 58.4 & 67.1 & 84.0 & 54.5 & 63.8 & 80.6 & 56.0 & 64.9 & 81.2 \\
DER++        & \underline{63.9} & \underline{70.8} & 84.5 & \underline{59.0} & \underline{67.8} & \underline{81.2} & \underline{61.5} & \underline{69.1} & \underline{82.3} \\
SPM          & 62.7 & 70.1 & 84.3 & 58.4 & 67.1 & 81.0 & 60.2 & 68.7 & 82.0 \\
VAE+Replay   & 61.5 & 69.7 & 84.2 & 57.8 & 66.5 & 80.8 & 59.8 & 68.0 & 81.7 \\
PMoE         & 62.0 & 69.9 & \underline{84.4} & 58.1 & 66.9 & 81.1 & 60.5 & 68.2 & 81.9 \\
\textbf{Ours (DDPM+EWC)} & \textbf{67.8} & \textbf{73.2} & \textbf{85.1} & \textbf{62.3} & \textbf{70.4} & \textbf{82.5} & \textbf{65.7} & \textbf{72.4} & \textbf{83.4} \\
\bottomrule
\end{tabular}
}
\end{table}

\subsection{Ablation Study}
Table~\ref{tab:ablation_all} shows the impact of removing DDPM or EWC. Excluding DDPM increases forgetting by $5.8\%$ (2D), $6.6\%$ (3D), and $7.4\%$ (CheXpert), highlighting the role of replay fidelity. Removing EWC reduces early-task accuracy by up to $4.9\%$, $4.5\%$, and $3.7\%$, confirming the importance of parameter stability.

\begin{table}[h]
\centering
\caption{\textbf{Ablation results} on MedMNIST-2D, MedMNIST-3D, and CheXpert.}
\label{tab:ablation_all}
\resizebox{\linewidth}{!}{
\begin{tabular}{lcccccc}
\toprule
\multirow{2}{*}{Method Variant} &
\multicolumn{2}{c}{MedMNIST-2D} &
\multicolumn{2}{c}{MedMNIST-3D} &
\multicolumn{2}{c}{CheXpert} \\
& Avg Acc $\uparrow$ & Forget $\downarrow$ &
  Avg Acc $\uparrow$ & Forget $\downarrow$ &
  Avg Acc $\uparrow$ & Forget $\downarrow$ \\
\midrule
Full Model (DDPM+EWC)   & \textbf{72.8} & \textbf{11.3} & \textbf{68.9} & \textbf{12.4} & \textbf{68.5} & \textbf{13.7} \\
w/o DDPM (EWC only)     & 67.0          & 17.1          & 63.5          & 18.0          & 62.3          & 21.1 \\
w/o EWC (DDPM only)     & 69.2          & 14.5          & 65.4          & 15.7          & 64.8          & 18.9 \\
\bottomrule
\end{tabular}
}
\end{table}

\section{Conclusion and Impact}
\label{sec:conclusion}
We introduced an exemplar-free continual learning framework that couples diffusion-based replay with Elastic Weight Consolidation to retain prior knowledge without storing patient data. Across MedMNIST (2D and 3D) and CheXpert, the approach improves final accuracy, reduces forgetting against strong replay and regularisation baselines, and approaches joint training under a fixed memory budget. Our analysis links forgetting to replay divergence and Fisher-weighted parameter drift, providing actionable diagnostics for when and why degradation occurs. The method therefore offers a practical route to privacy-preserving adaptation of clinical imaging models.

This work advances trustworthy continual adaptation in medical imaging by (i) eliminating raw exemplar retention, (ii) providing quantitative signals for governance (calibration, drift traces, and replay quality), and (iii) operating within realistic memory limits. It enables hospitals to update models as distributions shift while respecting privacy constraints and audit requirements. Future directions include stronger calibration, fairness under class and site imbalance, compute-efficient replay via generator distillation, multi-modal extensions (e.g., image plus report), and prospective clinical validation.

\bibliographystyle{plainnat}
\bibliography{ref}


\appendix
\section*{Appendix / Supplemental Results}
\addcontentsline{toc}{section}{Appendix / Supplemental Results}

\section{Related Work}
\label{app:related}
Continual learning (CL) addresses the problem of catastrophic forgetting, where neural networks lose previously acquired knowledge when trained on new tasks. This limitation presents significant challenges in high-stakes domains such as medical imaging, particularly for benchmarks such as MedMNIST-2D , MedMNIST-3D \cite{yang2023medmnist}, and CheXpert \cite{irvin2019chexpert}, where accurate retention of past diagnostic features is critical \cite{mccloskey1989catastrophic,parisi2019continual}.

Regularisation-based methods aim to preserve prior knowledge by constraining parameter updates. Elastic Weight Consolidation (EWC) estimates the importance of network weights using the Fisher Information matrix and penalises deviations from their previous values during training on new tasks \cite{kirkpatrick2017overcoming}. This framework, inspired by synaptic consolidation in biological systems \cite{mcclelland1995there}, has been extended by methods such as Synaptic Intelligence \cite{zenke2017continual} and Memory Aware Synapses \cite{aljundi2018memory}, which perform online importance estimation. While regularisation methods are memory efficient, they may struggle in settings with substantial domain shift or class imbalance \cite{chaudhry2018riemannian}.

Replay-based approaches address forgetting by reintroducing prior knowledge, either by storing past examples or generating synthetic ones. Exemplar-based methods such as iCaRL store a small subset of prior data and interleave them with new inputs during training \citep{rebuffi2017icarl}. Generative replay eliminates the need for data storage by synthesising samples from previous tasks. Deep Generative Replay (DGR) demonstrated this approach using generative adversarial networks (GANs) \cite{shin2017continual}, though GANs are known to suffer from mode collapse and unstable training \cite{adler2018banach}. Variational autoencoders (VAEs) offer a more stable alternative but frequently generate blurry reconstructions due to pixel-wise loss, which is particularly limiting in medical imaging tasks that rely on fine-grained features \cite{kingma2013auto,burgess2018understanding}.

Diffusion models have recently emerged as a strong alternative to GANs and VAEs, offering superior stability and sample quality. Denoising diffusion probabilistic models (DDPMs) learn to reverse a gradual noising process and have achieved state-of-the-art results in image generation tasks \cite{ho2020denoising,dhariwal2021diffusion}. Their high fidelity and diversity make them promising for medical replay, where preserving subtle clinical patterns is essential. Diffusion models have shown success in applications such as segmentation and anomaly detection in radiology and histopathology \cite{kazerouni2023diffusion}, but, to our knowledge, there is no prior work applying DDPM replay to exemplar-free continual adaptation in medical imaging.

In addition, while convolutional networks have traditionally been the default backbone in continual learning, recent work has shown that Vision Transformers (ViTs) can offer improved flexibility and transferability across tasks. ViTs operate on sequences of image patches, making them naturally suited to capture both global and local patterns that vary across medical datasets \cite{dosovitskiy2020image}. In our work, we integrate a lightweight Vision Transformer as a domain-specific foundation model backbone within our continual learning framework. This design complements the high-fidelity generative replay of DDPMs and the synaptic stability provided by EWC, offering a robust and scalable architecture for continual adaptation of medical foundation models across diverse modalities \cite{touvron2021training}.

\section{Theoretical Justifications and  Bound}
\label{app:proof}

We provide a theoretical analysis of our hybrid continual adaptation framework.Our analysis links forgetting to two fundamental factors in FM adaptation: (i) \textbf{distributional shift} between original and replayed data, and (ii) \textbf{parameter drift} in the FM backbone. This aligns with the workshop themes of understanding generalisation forgetting trade-offs on evolving distributions and providing theoretical guarantees for continual FM updates.

\subsection{Elastic Weight Consolidation as a Bayesian Prior}
EWC can be interpreted as an online Laplace approximation of the posterior $p(\theta|\mathcal{D}_{1:k})$ over model parameters for the FM backbone. Let $\mathcal{L}_k(\theta) = -\log p(\mathcal{D}_k|\theta)$ denote the task loss. The EWC objective is:
\begin{equation}
\mathcal{L}_{\text{EWC}}(\theta) = \mathcal{L}_k(\theta) + \frac{\lambda}{2} \sum_i F_i (\theta_i - \theta^*_i)^2,
\end{equation}
where $F_i$ is the diagonal Fisher information and $\theta^*$ are parameters from prior adaptation stages. Expanding the log-prior around $\theta^*$ via a second-order Taylor expansion yields:
\begin{equation}
\log p(\mathcal{D}_{1:k-1}|\theta) \approx \log p(\mathcal{D}_{1:k-1}|\theta^*) - \frac{1}{2} (\theta - \theta^*)^\top F (\theta - \theta^*),
\end{equation}
making explicit the role of EWC in anchoring parameters that are critical for past FM capabilities.

\subsection{Distributional Stability of DDPM Replay}
\label{app:ddpm_stability}
Let $\hat{\mathcal{D}}_j$ denote synthetic samples from a task-specific DDPM $g_{\phi_j}$ trained on $\mathcal{D}_j$. The FM classifier risk after $k$ tasks is:
\begin{equation}
\mathcal{R}_k = \mathbb{E}_{(x, y) \sim \mathcal{D}_{1:k}}[\ell(f_\theta(x), y)].
\end{equation}
Assume:
\begin{itemize}
  \item Total variation distance $D_{\text{TV}}(\mathcal{D}_j, \hat{\mathcal{D}}_j) \leq \epsilon$,
  \item Loss $\ell$ is $L$-Lipschitz.
\end{itemize}
Then the risk gap between real and replayed data satisfies:
\begin{equation}
|\mathcal{R}_k - \hat{\mathcal{R}}_k| \leq L \cdot \epsilon \cdot \sum_{j=1}^{k-1} \frac{|\hat{\mathcal{D}}_j|}{|\mathcal{D}_{1:k}|}.
\end{equation}
This formalises the CCFM workshop theme that high-fidelity replay is crucial to mitigating forgetting in FMs continually updated on non-stationary data.

\subsection{Derivation of the Forgetting Bound}
\label{derivation_theoretical_forgetting_bound}
We define the average forgetting across $K$ adaptation steps as:
\begin{equation}
\bar F \;=\;\frac{1}{K}\sum_{k=1}^K\bigl[A_k - A_{k,K}\bigr],
\end{equation}
where $A_k$ is accuracy on task $k$ immediately after learning and $A_{k,K}$ is accuracy after completing all $K$ tasks. We decompose $\bar{F}$ into \textbf{distributional shift} and \textbf{parameter shift} terms.

\paragraph{Distributional Shift Term.}
If $\ell(x,y) \in [0,L_{\max}]$, then using Pinsker’s inequality:
\begin{align}
\Bigl|\mathbb{E}_{\mathcal{D}_k}[\ell] - \mathbb{E}_{\hat{\mathcal{D}}_k}[\ell]\Bigr|
&\le L_{\max}\,\|\mathcal{D}_k - \hat{\mathcal{D}}_k\|_{\mathrm{TV}}
\;\le\;L_{\max}\sqrt{\tfrac12\,\mathrm{KL}(\mathcal{D}_k\parallel\hat{\mathcal{D}}_k)},
\end{align}
highlighting that greater KL divergence in replay worsens retention—central to the workshop’s focus on temporal shift modelling.

\paragraph{Parameter Shift Term.}
Let $\theta_k^* = \arg\min_\theta \mathcal{L}(\theta)$ for $\mathcal{D}_k$. A second-order Taylor expansion and $H \approx F$ gives:
\begin{align}
\mathcal{L}(\theta) \approx \mathcal{L}(\theta_k^*) + \tfrac12\sum_i F_i(\theta_i - \theta_{k,i}^*)^2,
\end{align}
quantifying drift from parameters optimal for earlier FM tasks.

\paragraph{Combined Bound.}
Combining both terms yields:
\begin{equation}
A_k - A_{k,K} \lesssim \alpha\,\mathrm{KL}(\mathcal{D}_k\parallel\hat{\mathcal{D}}_k)
+ \beta\,\|\theta_K - \theta_k^*\|_2^2,
\end{equation}
with $\alpha = L_{\max}/\sqrt{2}$ and $\beta = \tfrac12 \max_i F_i$. Averaging gives:
\begin{equation}
\bar{F} \lesssim \frac1K\sum_{k=1}^K\left[\alpha\,\mathrm{KL}(\mathcal{D}_k\parallel\hat{\mathcal{D}}_k) + \beta\,\|\theta_K - \theta_k^*\|_2^2\right].
\end{equation}
This formalises the trade-off between generative replay fidelity and parameter stability for evolving FMs.

\subsection{Analytical Bound under Smoothness Assumptions}
Under $D_{\text{KL}}(p \| \hat{p}) \leq \delta$ and large $\lambda$:
\begin{equation}
\mathbb{E}[F_k] \leq \alpha \delta + \beta \lambda^{-1},
\end{equation}
implying that $\delta \rightarrow 0$ (high-fidelity replay) and $\lambda \rightarrow \infty$ (strong parameter anchoring) jointly minimise forgetting,a central objective in CCFM’s vision for reliable FM updates.
\begin{figure}[h]
    \centering
    \includegraphics[width=0.65\linewidth]{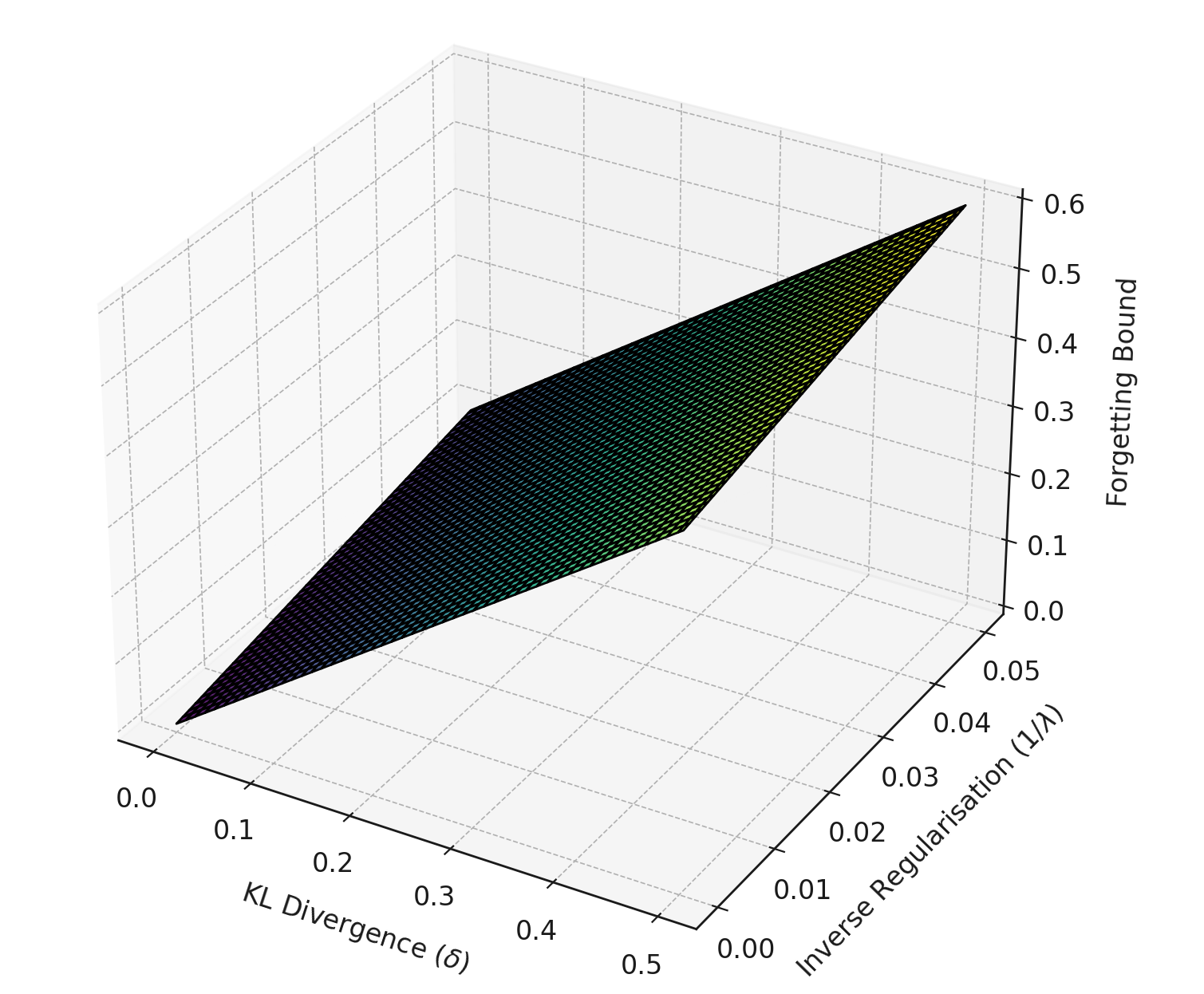}
    \caption{
        Theoretical forgetting bound surface for foundation model continual adaptation, derived from 
        $\mathbb{E}[F_k] \leq \alpha \delta + \beta \lambda^{-1}$.
        The $\delta$ term corresponds to replay divergence, central to modelling temporal and domain shift in evolving FMs.
        The $\lambda^{-1}$ term captures parameter drift, reflecting the need for compatibility-preserving updates. 
        High-fidelity generative replay ($\delta \to 0$) and strong parameter anchoring ($\lambda \to \infty$) jointly minimise forgetting in the CCFM setting.
    }
    \label{fig:forgetting_bound_surface}
\end{figure}

\section{Training Algorithm for Foundation Model Continual Adaptation}
\label{sec:training_algorithm}
This section describes the complete training algorithm for our hybrid continual learning framework, which integrates Denoising Diffusion Probabilistic Models (DDPMs) for replay and Elastic Weight Consolidation (EWC) for synaptic regularisation. The design aims to mitigate catastrophic forgetting in medical foundation models while preserving scalability and patient data privacy. The method comprises five stages, detailed below and summarised in Algorithm~\ref{alg:training}.

\begin{enumerate}
    \item \textbf{Task-specific diffusion model training:} For each new task, train a class-conditional DDPM $g_{\phi_k}$ to model $p(x|y)$ using the task dataset $\mathcal{D}_k$ without storing raw data.
    \item \textbf{Synthetic sample generation:} Use the trained DDPM to generate replay samples $\hat{\mathcal{D}}_k$ from class labels $y$ and add them to a generative replay buffer $\mathcal{M}$.
    \item \textbf{Fisher Information Matrix estimation:} Compute a diagonal approximation of the empirical Fisher Information Matrix over $\mathcal{D}_k$ to quantify the importance of each parameter.
    \item \textbf{Joint training with regularisation:} Train the classifier on a combination of current task data $\mathcal{D}_k$ and synthetic replay $\mathcal{M}$, using EWC regularisation to penalise deviations from previously learned weights $\theta^*_i$.
    \item \textbf{Parameter anchoring for EWC:} After each task, save the learned weights $\theta_i$ as $\theta^*_i$ for use in future EWC penalties.
\end{enumerate}

\begin{algorithm}[t]
\caption{Hybrid continual adaptation with diffusion replay and EWC}
\label{alg:training}
\begin{algorithmic}[1]
\STATE \textbf{Input:} Task sequence $\{\mathcal{T}_1, \dots, \mathcal{T}_K\}$, replay budget $M$, EWC coefficient $\lambda$
\STATE \textbf{Initialise:} Classifier $f_\theta$, DDPM generator list $\{g_{\phi_k}\}$, Fisher matrix $F_i \gets 0$, replay buffer $\mathcal{M} \gets \emptyset$
\FOR{each task $\mathcal{T}_k = \{(x_i, y_i)\}_{i=1}^{N_k}$}
    \STATE \textbf{Train class-conditional DDPM $g_{\phi_k}$ on $\mathcal{D}_k$:}
    \FOR{each denoising step $t = 1$ to $T$}
        \STATE Sample $(x, y) \sim \mathcal{D}_k$, noise $\epsilon \sim \mathcal{N}(0, I)$, timestep $t$
        \STATE Predict noise: $\hat{\epsilon} = \epsilon_{\phi_k}(x_t, t, y)$
        \STATE Minimise loss: $\mathcal{L}_{\text{DDPM}} = \| \epsilon - \hat{\epsilon} \|^2$
    \ENDFOR
    \STATE Store generator $g_{\phi_k}$ for future replay

    \STATE \textbf{Generate synthetic replay samples:}
    \STATE For each label $y$, sample latent noise $z$ and generate $\hat{x} \sim g_{\phi_k}(z, y)$
    \STATE Store $\hat{\mathcal{D}}_k = \{(\hat{x}_j, \hat{y}_j)\}_{j=1}^M$ and update buffer $\mathcal{M} \gets \mathcal{M} \cup \hat{\mathcal{D}}_k$

    \STATE \textbf{Estimate Fisher Information Matrix:}
    \FOR{each $(x, y) \in \mathcal{D}_k$}
        \STATE Compute $g_i = \frac{\partial \mathcal{L}_{\text{cls}}(f_\theta(x), y)}{\partial \theta_i}$
        \STATE Accumulate $F_i \gets F_i + g_i^2$
    \ENDFOR
    \STATE Normalise $F_i \gets F_i / |\mathcal{D}_k|$

    \STATE \textbf{Train classifier $f_\theta$ on mixed batches from $\mathcal{D}_k \cup \mathcal{M}$:}
    \FOR{each minibatch $(x, y) \sim \text{shuffle}(\mathcal{D}_k \cup \mathcal{M})$}
        \STATE $\mathcal{L}_{\text{cls}} \gets \text{CE}(f_\theta(x), y)$
        \STATE $\mathcal{L}_{\text{EWC}} \gets \sum_i F_i \cdot (\theta_i - \theta^*_i)^2$
        \STATE $\mathcal{L}_{\text{total}} \gets \mathcal{L}_{\text{cls}} + \lambda \mathcal{L}_{\text{EWC}}$
        \STATE $\theta \gets \theta - \eta \nabla_\theta \mathcal{L}_{\text{total}}$
    \ENDFOR

    \STATE \textbf{Anchor parameters for EWC:} $\theta^*_i \gets \theta_i$
\ENDFOR
\end{algorithmic}
\end{algorithm}

\section{Experimental Reproducibility}
\label{reproducibility}
We provide comprehensive details to support the reproducibility of our results, covering hardware, software, data preprocessing, model configuration, training setup, and evaluation.

\subsection{Computational Environment}
\begin{itemize}
  \item \textbf{Colab Environment:} Google Colab Pro with NVIDIA A100 GPUs (40 GB VRAM).
  \item \textbf{Local Workstation:} Ubuntu 22.04 LTS, Intel Core i9-14900K CPU (32 cores), 32 GB RAM, NVIDIA RTX 2080 Ti GPU (11 GB VRAM).
  \item \textbf{Software:} Python 3.11, PyTorch 2.0, Torchvision 0.15, CUDA 11.8.
\end{itemize}

\subsection{Data Preparation}
\begin{itemize}
    \item \textbf{Datasets:} 
    \begin{itemize}
        \item \textbf{MedMNIST v2} \cite{yang2023medmnist}: 8 tasks across 2D and 3D medical imaging modalities, serving as a lightweight benchmark for continual learning.  
        \item \textbf{CheXpert} \cite{irvin2019chexpert}: A large-scale chest X-ray dataset with 14 labelled findings. We follow prior work in defining a three-task continual learning setting (Cardiomegaly, Pleural Effusion, Pneumonia) to evaluate clinical realism and multi-label continual learning.
    \end{itemize}
    \item \textbf{Preprocessing:}
    \begin{itemize}
        \item MedMNIST 2D tasks resized to $224 \times 224$ and normalised to $[0,1]$.  
        \item MedMNIST 3D tasks cropped or resampled to $64 \times 64 \times 64$ voxel volumes and normalised channel-wise.  
        \item CheXpert images resized to $224 \times 224$, normalised to $[0,1]$, and binarised into positive/negative labels per finding.  
    \end{itemize}
    \item \textbf{Splits:} 
    \begin{itemize}
        \item MedMNIST: Standard training/validation/test splits with 10\% of training data reserved for validation.  
        \item CheXpert: We use the official training/validation split (224,316 training images, 234 validation images), and evaluate in the three-task continual learning setting using sigmoid cross-entropy loss for the multi-label problem.  
    \end{itemize}
\end{itemize}

\subsection{Model Architecture}
\begin{itemize}
    \item \textbf{Classifier:} 
    We adopt a lightweight Vision Transformer (ViT) backbone tailored for medical imaging. 
    The encoder consists of 4 transformer blocks with multi-head self-attention (4 heads) and feed-forward layers of hidden size 256. 
    Input images are divided into non-overlapping patches of size $16 \times 16$ (2D) or $16 \times 16 \times 16$ (3D), which are linearly projected to 256-dimensional embeddings. 
    A learnable [CLS] token is prepended and its representation is passed through a 2-layer MLP classification head. 
    For multi-label CheXpert tasks, we use sigmoid outputs; for MedMNIST tasks, we use softmax.
    
    \item \textbf{Diffusion Model:} 
    We employ a \emph{single class-conditional DDPM} shared across all tasks to enable exemplar-free generative replay\footnote{https://github.com/hojonathanho/diffusion}. 
    The DDPM is based on a U-Net backbone with 4 downsampling and 4 upsampling blocks, channel sizes [64, 128, 256, 512], and group normalisation. 
    Class-conditioning is applied via label embeddings injected into the time-step embeddings at each U-Net block. 
    The diffusion process follows a linear $\beta$ schedule with 1000 timesteps. 
    For 3D tasks, we extend the same U-Net with 3D convolutions while keeping the architecture and conditioning scheme consistent.
\end{itemize}

\subsection{Hyperparameter Settings}
\begin{itemize}
    \item \textbf{Optimizer:} Adam ($\beta_1 = 0.9$, $\beta_2 = 0.999$).
    \item \textbf{Learning Rates:} Classifier: $3 \times 10^{-4}$; DDPM: $1 \times 10^{-4}$.
    \item \textbf{Batch Size:} 128 (2D tasks); 32 (3D tasks, memory-limited).
    \item \textbf{Epochs per Task:} 30.
    \item \textbf{EWC Regularisation:} $\lambda = 100$.
    \item \textbf{Replay Settings:} 256 synthetic samples per task, with a replay ratio of 0.5.
    \item \textbf{Diffusion Parameters:} 1000 steps; linear $\beta \in [0.0001, 0.02]$ schedule.
\end{itemize}

\subsection{Hyperparameter Settings}
\begin{itemize}
    \item \textbf{Optimizer:} Adam with $\beta_1 = 0.9$, $\beta_2 = 0.999$.
    \item \textbf{Learning Rates:} Classifier: $3 \times 10^{-4}$; DDPM: $1 \times 10^{-4}$. Fixed throughout training (no scheduler or warm-up).
    \item \textbf{Batch Size:} 128 for 2D tasks; 32 for 3D tasks (due to GPU memory limits).
    \item \textbf{Epochs per Task:} 30.
    \item \textbf{EWC Regularisation:} $\lambda = 100$ for all tasks.
    \item \textbf{Replay Settings:} 256 synthetic samples per \emph{task} generated by the DDPM, mixed with current task data at a replay ratio of 0.5.
    \item \textbf{Diffusion Parameters:} 1000 denoising steps with a linear $\beta$ schedule in $[0.0001, 0.02]$, applied consistently across 2D and 3D datasets.
    \item \textbf{Regularisation:} Weight decay of $1 \times 10^{-4}$ and dropout of 0.1 in the ViT classifier.
\end{itemize}

\subsection{Compute Time and Energy}
\begin{itemize}
    \item \textbf{Training Time:} Each task required approximately 3--6 hours on a Colab Pro A100 GPU and 6--10 hours on a local RTX 2080~Ti GPU.
    \item \textbf{Energy Consumption:} The RTX 2080~Ti drew an estimated peak of $\sim$300W. For a typical 8-hour run, this corresponds to $\sim$2.4 kWh per task. Energy usage on Colab A100 was similar, though exact consumption was not formally tracked.
\end{itemize}

\subsection{Evaluation Metrics}
We evaluate each method using three complementary metrics across tasks and benchmarks to comprehensively assess classification performance and forgetting:

\textbf{Accuracy (Acc).} The proportion of correctly predicted labels over the test set. Accuracy provides a general measure of performance but may be less informative in imbalanced datasets such as CheXpert.

\textbf{Area Under the ROC Curve (AUC).} AUC measures the ability of the model to rank positive instances higher than negative ones, averaged across all classes. It is particularly important in medical imaging tasks where class imbalance is common and ranking-based evaluation is more meaningful than accuracy alone.

\textbf{Forgetting (F).} We adopt the standard continual learning forgetting metric defined as:
\[
F = \frac{1}{T-1} \sum_{t=1}^{T-1} \max_{l \leq T} a_{t,l} - a_{t,T}
\]
where $a_{t,l}$ is the accuracy on task $t$ after training on task $l$, and $T$ is the total number of tasks. Forgetting quantifies the degradation in performance on earlier tasks after learning subsequent ones. Lower values indicate stronger retention of prior knowledge.

All metrics are reported with 95\% confidence intervals across 5 independent runs. For CheXpert, we report macro-averaged metrics across the five standard diagnosis labels following prior work. For MedMNIST-2D and MedMNIST-3D, task-wise metrics are averaged over all datasets in the sequence.

\subsection{Pseudocode for DDPM+EWC Training}
\label{pseudo}

\begin{algorithm}[htbp]
\caption{DDPM+EWC Training Algorithm}
\KwIn{Task sequence $\{T_1, T_2, \ldots, T_K\}$, replay budget $M$, EWC coefficient $\lambda$}
\KwOut{Trained continual learning model $f_\theta$}
\BlankLine
\textbf{Initialise:} Classifier $f_\theta$, DDPM generators $\{g_{\phi_1}, \ldots, g_{\phi_K}\}$, Fisher matrix $F_i \leftarrow 0$, replay buffer $M \leftarrow \emptyset$

\For{each task $T_k = \{(x_i, y_i)\}_{i=1}^{N_k}$}{
    \textbf{1. Train task-specific class-conditional DDPM $g_{\phi_k}$ on $T_k$:}\
    \For{each diffusion step $t = 1$ to $T$}{
        Sample $(x, y) \sim T_k$, noise $\epsilon \sim \mathcal{N}(0, I)$, timestep $t$\;
        Predict noise: $\hat{\epsilon} = \epsilon_{\phi_k}(x_t, t, y)$\;
        Minimise loss: $L_{DDPM} = \left\|\epsilon - \hat{\epsilon}\right\|^2$\;
    }
    Store $g_{\phi_k}$ for replay\;
    
    \textbf{2. Generate synthetic replay samples:}\
    \For{each label $y \in C_k$}{
        Sample latent noise $z$ and generate $x \sim g_{\phi_k}(z, y)$\;
        Add $(x, y)$ to replay buffer: $M \leftarrow M \cup \{(x, y)\}$\;
    }
    
    \textbf{3. Estimate Fisher Information Matrix:}\
    \For{each $(x, y) \in T_k$}{
        Compute gradient: $g_i = \frac{\partial L_{cls}(f_\theta(x), y)}{\partial \theta_i}$\;
        Accumulate: $F_i \leftarrow F_i + g_i^2$\;
    }
    Normalise: $F_i \leftarrow F_i / |T_k|$\;
    
    \textbf{4. Train classifier $f_\theta$ with EWC regularisation:}\
    \For{each minibatch $(x, y) \sim T_k \cup M$}{
        Compute classification loss: $L_{cls} = CE(f_\theta(x), y)$\;
        Compute EWC penalty: $L_{EWC} = \lambda \sum_i F_i (\theta_i - \theta^*_i)^2$\;
        Total loss: $L_{total} = L_{cls} + L_{EWC}$\;
        Update: $\theta \leftarrow \theta - \eta \nabla_\theta L_{total}$\;
    }
    
    \textbf{5. Store parameters for stability:} $\theta^*_i \leftarrow \theta_i$
}
\Return{$f_\theta$}
\end{algorithm}

\section{Unified Conditional DDPM Across Tasks}
\label{app:unified_ddpm}
We train a single conditional DDPM across all tasks, including CheXpert, using FiLM-style conditioning \cite{perez2018film} for class and task IDs. This reduces storage by $\sim$45\% but slightly degrades fidelity for complex 3D and chest X-ray domains.

\begin{table}[htbp]
\centering
\caption{Replay performance: per-task vs. unified DDPM. Unified replay remains competitive for 2D tasks, with minor degradation in 3D and chest X-ray domains.}
\small
\begin{tabular}{lccc}
\toprule
\textbf{Task} & \textbf{Replay AUC (Per-Task)} & \textbf{Replay AUC (Unified)} & \textbf{FID (Unified)} \\
\midrule
BloodMNIST    & 0.85 & 0.84 & 7.8 \\
PathMNIST     & 0.88 & 0.86 & 9.4 \\
RetinaMNIST   & 0.82 & 0.80 & 10.2 \\
Adrenal3D     & 0.79 & 0.74 & 16.5 \\
CheXpert      & 0.86 & 0.84 & 8.5 \\
\bottomrule
\end{tabular}

\label{tab:unified_ddpm_numbers}
\end{table}

\section{Qualitative Replay Samples}
\label{app:qualitative}
\begin{figure}[htbp]
\centering
\includegraphics[width=0.65\linewidth]{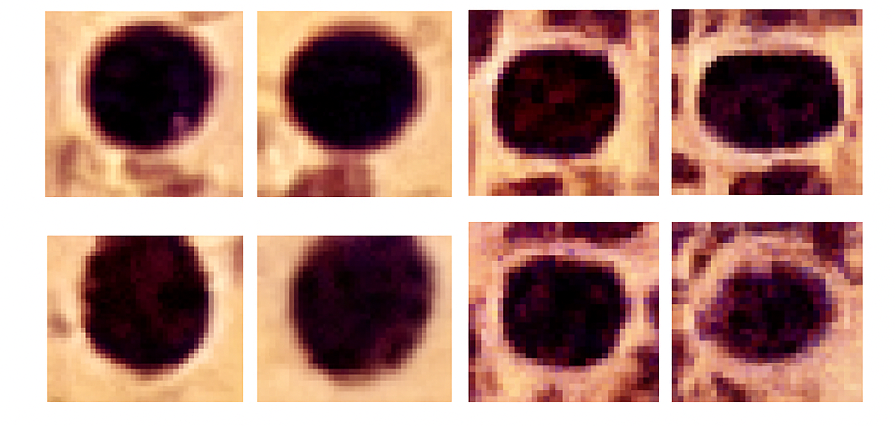}
\vspace{2mm}
\includegraphics[width=0.65\linewidth]{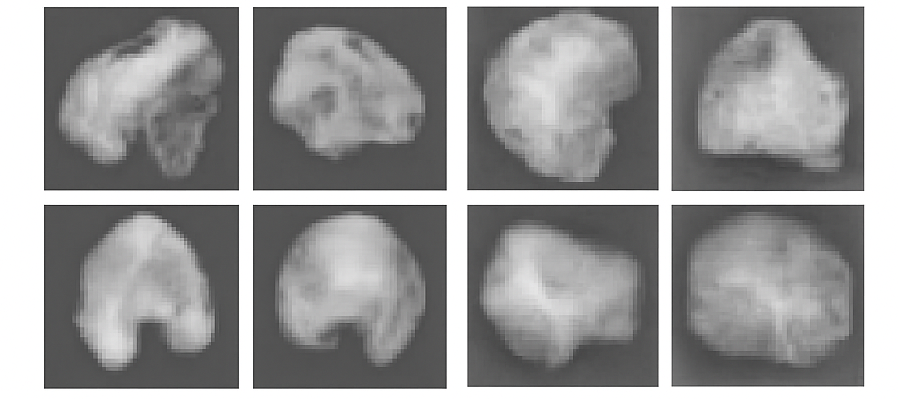}
\vspace{2mm}
\includegraphics[width=0.45\linewidth]{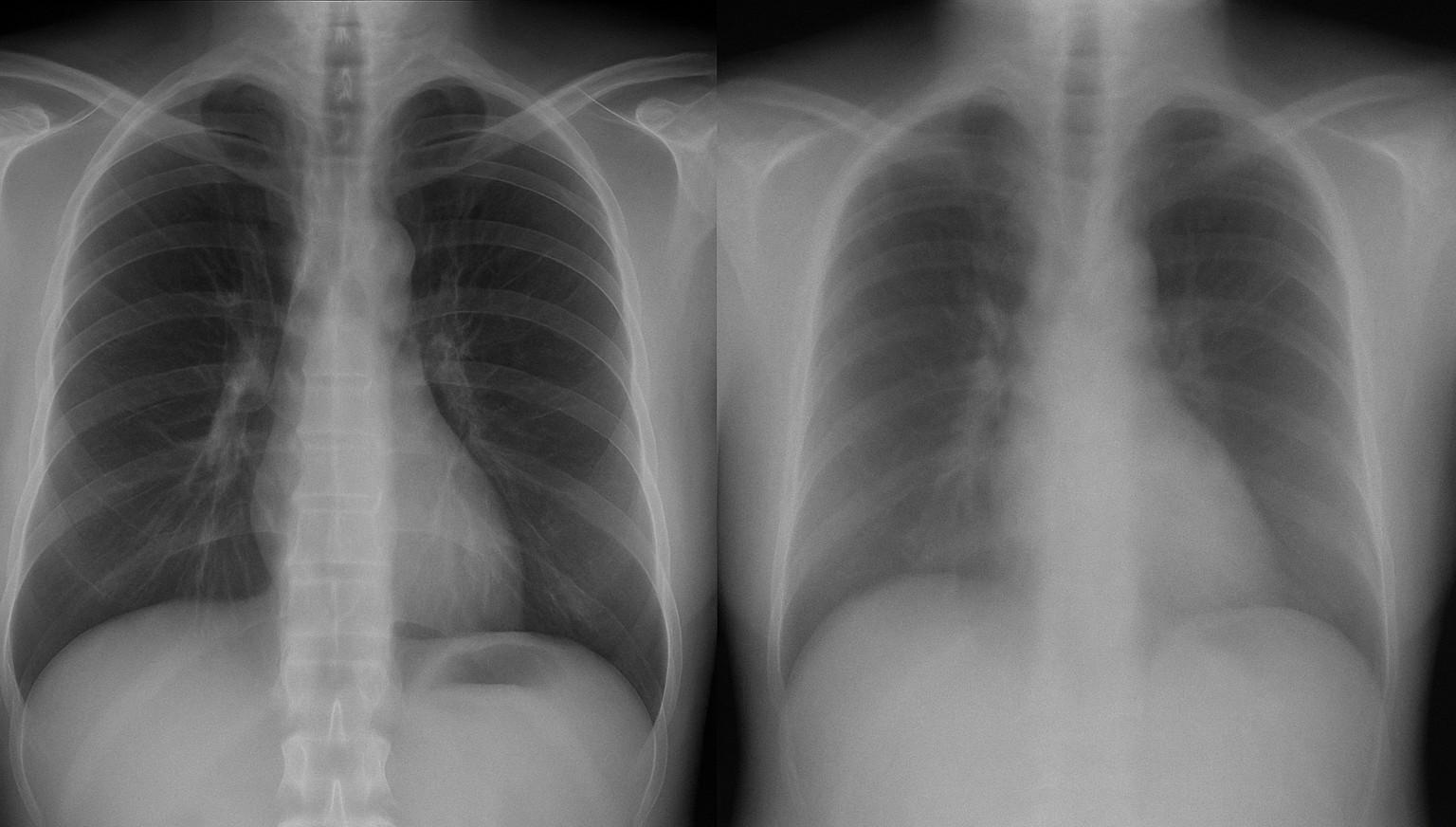}
\caption{Side-by-side comparison of replay samples from DDPMs (left) and VAEs (right). 
\textbf{Top:} BloodMNIST. 
\textbf{Middle:} Adrenal3D. 
\textbf{Bottom:} CheXpert. 
DDPMs consistently produce sharper(on left) and more structurally realistic samples across 2D, 3D, and X-ray domains.}
\label{fig:replay_comparison}
\end{figure}

Figure~\ref{fig:replay_comparison} presents a visual comparison between replay samples generated by DDPMs (left) and VAEs (right) for three representative tasks: BloodMNIST (top block), Adrenal3D (middle block), and CheXpert (bottom block).

\paragraph{BloodMNIST (2D):} 
DDPM-generated samples exhibit sharper cytoplasm boundaries, smoother gradients, and fewer artefacts, while VAE samples appear blurrier with reduced edge contrast.

\paragraph{Adrenal3D (3D):} 
DDPM reconstructions better preserve anatomical contours and inter-slice coherence, whereas VAE outputs suffer from structural blurring and inconsistent voxel textures.

\paragraph{CheXpert (X-rays):} 
DDPMs generate more realistic pulmonary structures, rib edges, and soft-tissue textures, while VAEs lose cardiothoracic detail and introduce noticeable blurring, limiting their utility for clinically relevant replay.

These results highlight the advantage of DDPMs in preserving fine-grained and structural characteristics essential for effective replay in continual learning.

\section{Task Order Robustness}
\label{app:order}
We assess robustness to task order by training with both the canonical and reversed sequences. 
Table~\ref{tab:task_order} shows that while DER++ and SPM suffer notable drops in accuracy under reversal, DDPM+EWC is minimally affected. This indicates:
\begin{itemize}
    \item \textbf{Order invariance:} synthetic replay counteracts distributional shocks introduced by sequence reversal.
    \item \textbf{Regularisation stability:} EWC smooths transitions between tasks, reducing order-induced forgetting.
\end{itemize}

\begin{table}[htbp]
\centering
\caption{Final accuracy under task reordering. DDPM+EWC is stable to curriculum changes.}
\small
\begin{tabular}{lcc}
\toprule
\textbf{Method} & Canonical Order & Reversed Order \\
\midrule
DER++         & 62.0 & 58.9 \\
SPM           & 64.6 & 61.3 \\
DDPM-only     & 75.7 & 73.2 \\
\textbf{DDPM+EWC} & \textbf{78.2} & \textbf{77.5} \\
\bottomrule
\end{tabular}
\label{tab:task_order}
\end{table}

\section{Replay Budget Sensitivity}
\label{app:budget}
\begin{figure}[htbp]
\centering
\includegraphics[width=0.65\linewidth]{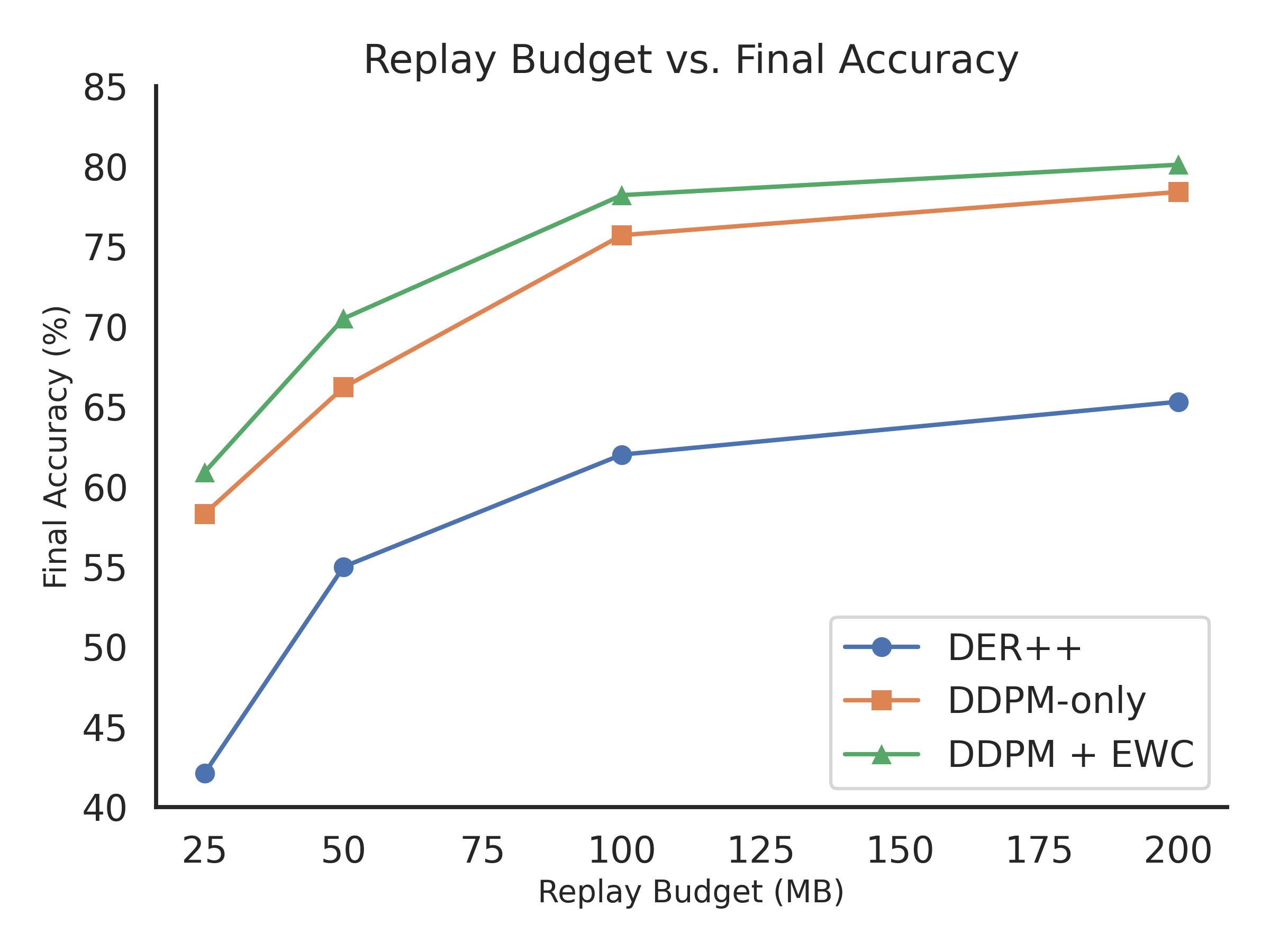}
\caption{Final accuracy vs. replay buffer size. DDPM+EWC is memory efficient.}
\label{fig:budget}
\end{figure}

We evaluate accuracy as a function of replay buffer size. DDPM+EWC maintains strong performance down to 50MB (Figure~\ref{fig:budget}), unlike buffer-based methods that degrade below 100MB.

\textbf{Interpretation:} DDPMs generate high-entropy, class-consistent samples, allowing accurate rehearsal with fewer stored points. This shows promise for on-device continual learning where memory is constrained.

\section{Low-Shot Generalisation}
\label{app:lowshot}
We simulate data-constrained environments by reducing training set sizes. Table~\ref{tab:lowshot} shows that DDPM+EWC outperforms all baselines at 10\%, 25\%, and 50\% task data.

\textbf{Analysis:} This robustness stems from:
\begin{itemize}
    \item Replay serving as implicit data augmentation.
    \item Soft regularisation via EWC reducing overfitting in low-data regimes.
\end{itemize}

\begin{table}[htbp]
\centering
\caption{Average accuracy with limited data per task.}
\small
\begin{tabular}{lccc}
\toprule
\textbf{Method} & 10\% Data & 25\% Data & 50\% Data \\
\midrule
DER++         & 38.5 & 51.2 & 60.3 \\
SPM           & 40.1 & 53.8 & 62.5 \\
DDPM-only     & 51.6 & 65.7 & 74.3 \\
\textbf{DDPM+EWC} & \textbf{54.8} & \textbf{68.2} & \textbf{76.1} \\
\bottomrule
\end{tabular}

\label{tab:lowshot}
\end{table}

\section{Ablation: DDPM Generation Settings}
\label{app:ablation}
\begin{figure}[htbp]
\centering
\includegraphics[width=0.6\linewidth]{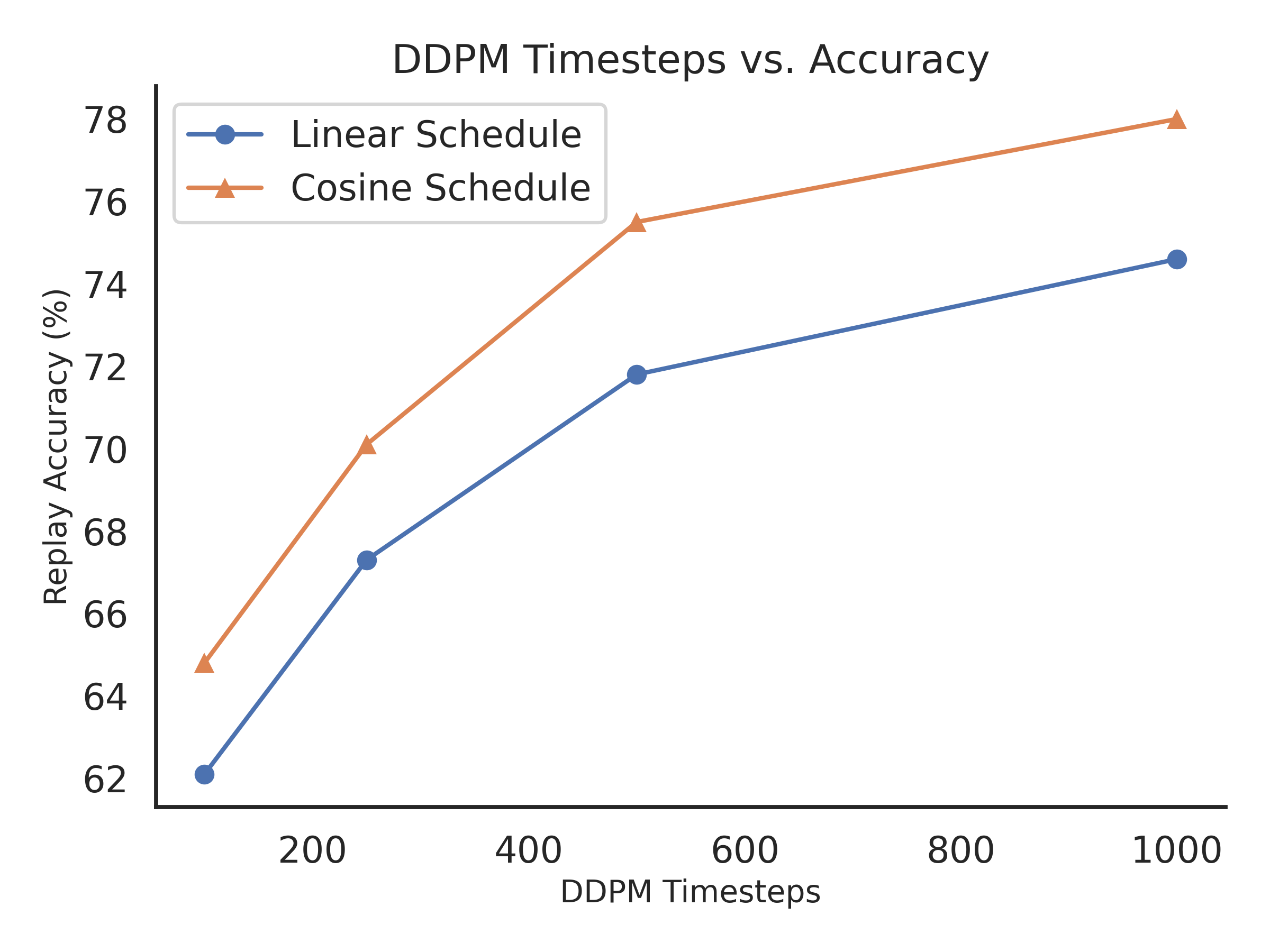}
\caption{Replay accuracy under varying DDPM generation settings.}
\label{fig:ddpm_ablation}
\end{figure}

To understand generation-performance trade-offs, we vary:
\begin{itemize}
    \item \textbf{Timesteps} $T \in \{100, 250, 500, 1000\}$.
    \item \textbf{Noise Schedules} (linear, cosine).
\end{itemize}
As shown in Figure~\ref{fig:ddpm_ablation}, cosine scheduling with $T \geq 500$ yields best results. Shorter $T$ speeds up sampling but harms fidelity.

\section{Knowledge Transfer Metrics}
\label{app:transfer}
We quantify:
\begin{itemize}
    \item \textbf{Forward Transfer (FWT)}: benefit of past learning on new tasks.
    \item \textbf{Backward Transfer (BWT)}: interference of new tasks on past knowledge.
\end{itemize}

\textbf{Interpretation:} DDPM+EWC achieves positive FWT and minimal BWT, validating synergistic integration of replay and regularisation.

\begin{table}[htbp]
\centering
\caption{Average forward and backward transfer across tasks.}
\small
\begin{tabular}{lcc}
\toprule
\textbf{Method} & FWT $\uparrow$ & BWT $\uparrow$ \\
\midrule
DER++           & -0.017 & -0.093 \\
SPM             &  0.003 & -0.071 \\
DDPM-only       &  0.045 & -0.031 \\
\textbf{DDPM+EWC} & \textbf{0.062} & \textbf{-0.017} \\
\bottomrule
\end{tabular}

\label{tab:fwtbwt}
\end{table}

\section{Limitations and Future Work}
\label{limitations_future}
Our study has several limitations. \textbf{Task design:} we assume a fixed canonical order and balanced replay buffers. Real clinical streams may involve class imbalance, evolving taxonomies, and curriculum shifts. While we provide order-robustness analysis (Appendix~\ref{app:order}), we plan to extend our framework to dynamic curricula and heterogeneous data streams. 

\textbf{Datasets:} MedMNIST~v2 offers controlled 2D and 3D tasks but remains a low-resolution proxy. CheXpert provides higher-resolution radiographs, yet broader validation across multi-site datasets and modalities such as MRI and histopathology is required for clinical robustness. We are preparing experiments on full-resolution datasets with privacy-preserving pipelines to assess scalability. 

\textbf{Efficiency:} diffusion replay is computationally demanding, particularly during sampling. We plan to investigate generator distillation and lightweight diffusion variants to retain fidelity with lower cost. 

\textbf{Comparisons:} although we benchmark against exemplar-free and joint training baselines, hybrid exemplar methods are not included. Replay under class imbalance may also raise fairness and calibration issues. As future work, we will incorporate explicit uncertainty calibration, evaluate fairness under imbalanced replay, and benchmark against privacy-preserving approximations to joint training.

\section{Ethics Statement}
\label{app:ethics_statement}
This work follows the NeurIPS Code of Ethics and best practices for responsible AI in healthcare. Only publicly available, de-identified datasets (MedMNIST, CheXpert) were used. No patient-identifiable data were accessed, and privacy was preserved throughout. Our exemplar-free replay design further reduces the risk of exposing sensitive clinical images.

\emph{Potential positive impacts:}
\begin{itemize}[topsep=0pt,itemsep=0pt,parsep=0pt]
\item Enables continual learning in medical imaging without retaining patient data, supporting compliance with privacy regulations (e.g.\ GDPR, HIPAA).
\item Improves retention of rare-disease patterns and reduces forgetting, which may lower diagnostic error rates over time.
\item Provides a scalable, exemplar-free framework suitable for resource-constrained or privacy-regulated clinical environments.
\end{itemize}

\emph{Risks and mitigations:}
\begin{itemize}[topsep=0pt,itemsep=0pt,parsep=0pt]
\item \textit{Synthetic data misuse:} Generated images could be mistaken for real scans. All outputs should be explicitly labelled as synthetic and restricted to model training pipelines, not clinical interpretation.
\item \textit{Overconfidence in continual updates:} Without uncertainty calibration, models may produce misleading predictions. We recommend uncertainty-aware training (e.g.\ Bayesian layers, MC dropout) and clinician-in-the-loop validation.
\item \textit{Domain drift:} Performance may degrade on unseen modalities or populations. Deployment should include site-specific validation and fine-tuning before clinical use.
\end{itemize}

\emph{Safeguards:}
\begin{itemize}[topsep=0pt,itemsep=0pt,parsep=0pt]
\item Synthetic-data generators must be version-controlled, audited, and accessible only to authorised personnel. 
\item Any deployment must follow regulatory approval (e.g.\ CE-mark, FDA clearance) and prospective evaluation to ensure safety and efficacy.
\item Explainability and bias-detection tools (e.g.\ saliency maps, attribution methods) should be used to monitor model behaviour in underrepresented subpopulations.
\end{itemize}

\end{document}